\documentclass[letterpaper,journal]{IEEEtran}

\usepackage[T1]{fontenc}
\usepackage[utf8]{inputenc}
\usepackage{amsmath,amssymb,amsfonts}
\usepackage{graphicx,array,booktabs,tabularx,multirow}
\usepackage{xcolor,cite,url,stfloats,balance,microtype}
\usepackage[hidelinks]{hyperref}
\usepackage{makecell}
\graphicspath{{figures/}}

\newcommand{\clip}{\operatorname{clip}}
\newcommand{\warp}{E_{\mathrm{warp}}}
\newcommand{\ind}{\mathbf{1}}
\newcommand{\R}{\mathbb{R}}
\newcommand{\nrm}[1]{\left\lVert #1\right\rVert}
\newcolumntype{Y}{>{\raggedright\arraybackslash}X}

\begin{document}
\title{IGSW: Multi-Weather Urban Scene Reconstruction\\with Intrinsic-Decoupled Gaussian Splatting}
\author{Zhongquan~Wang,
Long~Cao,
Peijia~Xie,
Senlei~Ma,
Jie~Li,
and~Guofa~Li%
\thanks{This work was supported by the National Natural Science Foundation of China under Grant 52272421. (Corresponding author: Guofa Li.)}%
\thanks{The authors are with the College of Mechanical and Vehicle
Engineering, Chongqing University, Chongqing 400044, China
(e-mail: wangzhongquan@stu.cqu.edu.cn;
caolong@stu.cqu.edu.cn;
xiepeijia@stu.cqu.edu.cn;
masenlei026@stu.cqu.edu.cn;
jieli@cqu.edu.cn;
liguofa@cqu.edu.cn).}%
}
\maketitle

\begin{abstract}
High-fidelity weather editing of 4D scenes can support end-to-end training and closed-loop simulation for autonomous driving. Combining generative image editing with dynamic Gaussian representations enables multi-weather scene reconstruction, but weather-edited images often contain structural distortions. Directly fitting their appearance may propagate errors from teacher images into the reconstructed 4D scene, compromising the original scene structure and temporal continuity. To address these challenges, we propose IGSW, a two-stage framework that combines intrinsic-decoupled 4D Gaussian scene modeling with structure-preserving weather refinement. In Stage~I, background Gaussian appearance is decomposed into weather-shared albedo and normals and weather-specific irradiance and residuals. Weather teacher images generated by a diffusion-based image editing model and intrinsic maps estimated by a diffusion-based inverse renderer provide feature-level soft supervision for stable structure and illumination changes. In Stage~II, a structure-preservation gate combines semantic conflicts with cross-weather intrinsic discrepancies to suppress unreliable teacher regions. Per-Gaussian weather transport variables and target-weather branches then strengthen the desired weather appearance over the intrinsic-decoupled scene. Experiments on nuScenes and Waymo show that IGSW achieves the highest content similarity and semantic consistency and the lowest temporal error among the compared methods, while retaining pronounced weather-dependent illumination and atmosphere. The project page is available at \url{https://akaxi6.github.io/IGSW/}.
\end{abstract}


\begin{figure*}[!t]
    \centering
    \includegraphics[width=\textwidth]{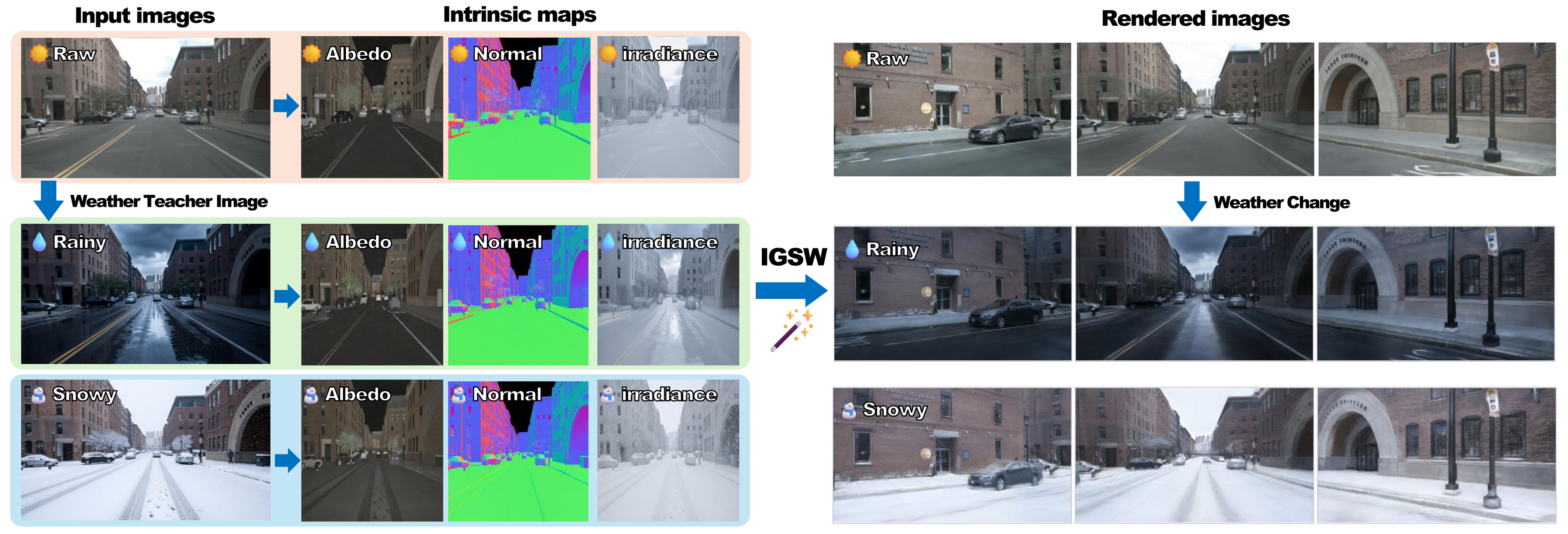}
    \caption{Teaser of IGSW. Given a driving-scene sequence, target-weather teacher images, and intrinsic maps (albedo, normal, and irradiance) estimated through inverse rendering, IGSW reconstructs a dynamic multi-weather urban scene in two stages: intrinsic-decoupled Gaussian appearance modeling and intrinsic-preserving weather refinement. The resulting Gaussian scene renders target-weather views while preserving the spatial structure of the original observations.}
    \label{fig:teaser}
\end{figure*}

\section{Introduction}
\label{sec:intro}
\IEEEPARstart{W}{eather} editing can extend captured driving scenes to conditions that are difficult to record with matched viewpoints and traffic states. Rain changes illumination and road reflections, whereas snow changes surface appearance and contrast. These effects should not alter the spatial layout or motion of roads, buildings, and traffic participants. Weather editing therefore requires more than realistic individual frames: it must construct a unified scene representation whose weather appearance remains consistent across views and time.

3D Gaussian Splatting~\cite{r1} and dynamic urban scene representations~\cite{r3,r4} provide a foundation for efficient reconstruction and rendering. Text-guided image editors such as Qwen-Image~\cite{r18} can further synthesize target-weather observations beyond the captured condition. WeatherEdit~\cite{r6} and WeatherCity~\cite{r5} combine such appearance priors with scene-level modeling, extending weather synthesis from independent image editing to three-dimensional scenes. Nevertheless, a unified representation alone does not distinguish valid weather changes from erroneous structural changes in edited images.

Strengthening weather appearance can also amplify implausible modifications to scene content. Generated clouds may establish a rainy atmosphere while occluding a building that should remain visible; convincing reflections may be accompanied by inconsistent vehicle details. Strongly anchoring optimization to the original RGB observations can suppress these errors, but may also weaken the intended weather effects. Effective weather editing must therefore answer two coupled questions: which scene factors should change with weather, and which teacher regions provide reliable supervision.

Intrinsic decomposition provides a basis for the first distinction. Albedo and surface normals offer material and structural cues, whereas irradiance describes illumination variation. Generative inverse rendering, as used by IntrinsicWeather~\cite{r19}, estimates these factors from original and weather-edited images. Independently estimated intrinsic maps, however, can contain local errors and cross-view inconsistencies.

We introduce \textbf{IGSW (Intrinsic Gaussian Splatting Weather)}, a two-stage framework for intrinsic-decoupled 4D Gaussian modeling and structure-preserving weather refinement. Stage~I decomposes background Gaussian appearance into weather-shared albedo and normals and weather-specific irradiance and residuals. Intrinsic feature supervision anchors shared content to the original scene and guides the weather branches toward the target illumination change. Stable content and weather variation thus assume distinct modeling roles within one renderable Gaussian scene.

Intrinsic decoupling does not imply that every teacher change is valid, because weather-specific branches may still reproduce structural errors. Stage~II addresses this issue through reliability-aware refinement. A structure-preservation gate combines semantic conflicts with differences between the original and edited albedo/normal estimates to identify unreliable teacher regions. Reliable regions retain teacher appearance supervision, whereas conflicting regions use weather-styled targets that preserve original spatial content. With geometry and the shared intrinsic representation fixed, per-Gaussian weather transport variables and target-weather branches refine illumination and residual appearance. Experiments on nuScenes and Waymo show that, among the compared scene-editing baselines, IGSW improves content preservation, semantic consistency, and temporal stability while retaining pronounced weather-dependent illumination and atmosphere. These properties support multi-weather urban reconstruction and subsequent data generation for autonomous-driving simulation.

Our main contributions are as follows:
\begin{itemize}
    \item We propose an intrinsic-decoupled 4D Gaussian weather representation. Gaussian appearance is separated into weather-shared albedo/normals and weather-specific irradiance/residuals, explicitly assigning stable scene content and weather variation different representational roles. Then we incorporate intrinsic information to guide the iterative optimization of the decoupled Gaussian representation, promoting material and structural consistency while modeling weather-induced illumination changes.
    \item We introduce an intrinsic-preserving Gaussian weather refinement method. A structure-preservation gate combines semantic conflicts with cross-weather albedo and normal discrepancies to mitigate the propagation of structural errors from weather-edited teacher images into the 3D scene. With geometry and shared intrinsic representations fixed, per-Gaussian weather transport variables and weather-specific branch optimization refine irradiance and residual responses, strengthening target-weather appearance while preserving the original scene structure.
    \item We validate IGSW on nuScenes and Waymo. Full-sequence rain and snow evaluations, visual comparisons, and ablation studies demonstrate that IGSW balances target-weather expression, preservation of original scene content and semantic structure, and temporal continuity, supporting multi-weather scene simulation and data generation for autonomous driving.
\end{itemize}

\section{Related Work}

\subsection{Image and Video Weather Editing}
Weather generation and restoration have long supported robustness research for autonomous driving. Early directions include GAN-based weather transfer~\cite{r11}, extreme-weather synthesis with ClimateNeRF~\cite{r10}, and multi-weather restoration~\cite{r12}. Diffusion and text-guided editors, including InstructPix2Pix~\cite{r13}, ControlNet~\cite{r14}, Prompt-to-Prompt~\cite{r15}, and Qwen-Image~\cite{r18}, provide more flexible control over image-level appearance. Video editors such as TokenFlow~\cite{r16} and FRESCO~\cite{r17} propagate correspondence to reduce frame-wise inconsistency. FluencyVE~\cite{cai2026fluencyve} uses temporal-aware Mamba and bypass attention for efficient video editing. Qian et al.~\cite{qian2026semantic} combine semantic appearance guidance, physical particle simulation, and geometric conditioning for weather video synthesis. IGSW distills weather edits into a dynamic 4D Gaussian scene while filtering teacher-induced content errors.

\begin{table}[!t]
\centering
\caption{Method categories and representation capabilities.}
\label{tab:methods}
\scriptsize
\setlength{\tabcolsep}{2.5pt}
\renewcommand{\arraystretch}{1.15}

\resizebox{\columnwidth}{!}{%
\begin{tabular}{@{}lclccc@{}}
\toprule
\makecell[l]{Method} &
\multicolumn{1}{l}{\makecell[l]{Venue}} &
\makecell[l]{Type} &
\makecell[c]{Intrinsic\\decomp} &
\makecell[c]{3D/4D\\scene} &
\makecell[c]{Temporal\\consist.} \\
\midrule

Qwen-Image
& arXiv 2025
& Frame editing
& $\times$ & $\times$ & $\times$ \\

FRESCO
& CVPR 2024
& Video editing
& $\times$ & $\times$ & $\checkmark$ \\

IntrinsicWeather
& CVPR 2026
& Intrinsic image editing
& $\checkmark$ & $\times$ & $\times$ \\

WeatherEdit
& AAAI 2026
& 3D/4D weather
& $\times$ & $\checkmark$ & $\checkmark$ \\

WeatherCity
& CVPR 2026
& 4D weather
& $\times$ & $\checkmark$ & $\checkmark$ \\

IGSW (Ours)
& --
& Intrinsic 4DGS
& $\checkmark$ & $\checkmark$ & $\checkmark$ \\

\bottomrule
\end{tabular}%
}
\end{table}

\subsection{Intrinsic Decomposition and Generative Inverse Rendering}
Intrinsic decomposition separates material, geometry, and illumination factors to provide a structured representation of scene appearance. Albedo describes the underlying material color, normals characterize surface orientation, and irradiance captures lighting variation, providing complementary cues for appearance editing and relighting. Yang et al.~\cite{yang2025relighting} incorporate intrinsic decomposition into a two-stage single-image relighting network. Intrinsic Image Diffusion~\cite{r20}, IntrinsicEdit~\cite{r21}, RGB$\leftrightarrow$X~\cite{r22}, Uni-Renderer~\cite{r23}, and DiffusionRenderer~\cite{r24} use generative priors to estimate or manipulate intrinsic properties. IntrinsicReal~\cite{wei2026intrinsicreal} adapts a diffusion-based albedo estimator to real objects using confidence- and ranking-based pseudo-label selection. IntrinsicWeather~\cite{r19} extends this reasoning to weather editing through albedo, normals, and irradiance, but remains an image-space method. IGSW instead models shared and weather-specific intrinsic factors within a dynamic Gaussian scene, supporting weather-dependent appearance changes while preserving stable scene content.

\subsection{Three-Dimensional Weather Scene Modeling}
NeRF~\cite{r2} and 3D Gaussian Splatting~\cite{r1} reconstruct and render multi-view scenes through implicit radiance fields and explicit Gaussian primitives, respectively. Street Gaussians~\cite{r3} and DrivingGaussian~\cite{r4} extend reconstruction to dynamic traffic participants. TrAME~\cite{luo2025trame} couples progressive view editing with Gaussian updates to improve multi-view consistency. Cao et al.~\cite{cao2026supportaware} improve cross-view consistency through support-aware appearance residual distillation. These representations provide the basis for scene-level weather editing, but observations from one condition do not determine how the same scene should appear under another. WeatherGS~\cite{qian2025weathergs} removes weather artifacts and masks lens occlusions for Gaussian reconstruction, whereas WeatherDrive~\cite{hua2025weatherdrive} studies multi-view street-scene generation under multiple weather conditions. ClimateNeRF~\cite{r10}, WeatherEdit~\cite{r6}, Weather-Magician~\cite{r7}, and WeatherCity~\cite{r5} incorporate weather variation into three-dimensional or dynamic scenes, whereas RainyGS~\cite{r8} and Let It Snow~\cite{r9} model specific rain or snow effects. Even with fixed Gaussian positions, fitting edited images may reproduce false occlusions and content changes. IGSW addresses these errors through intrinsic factorization and reliability-aware refinement.

\section{Method}
\label{sec:method}

\begin{figure*}[!t]
    \centering
    \includegraphics[width=\textwidth]{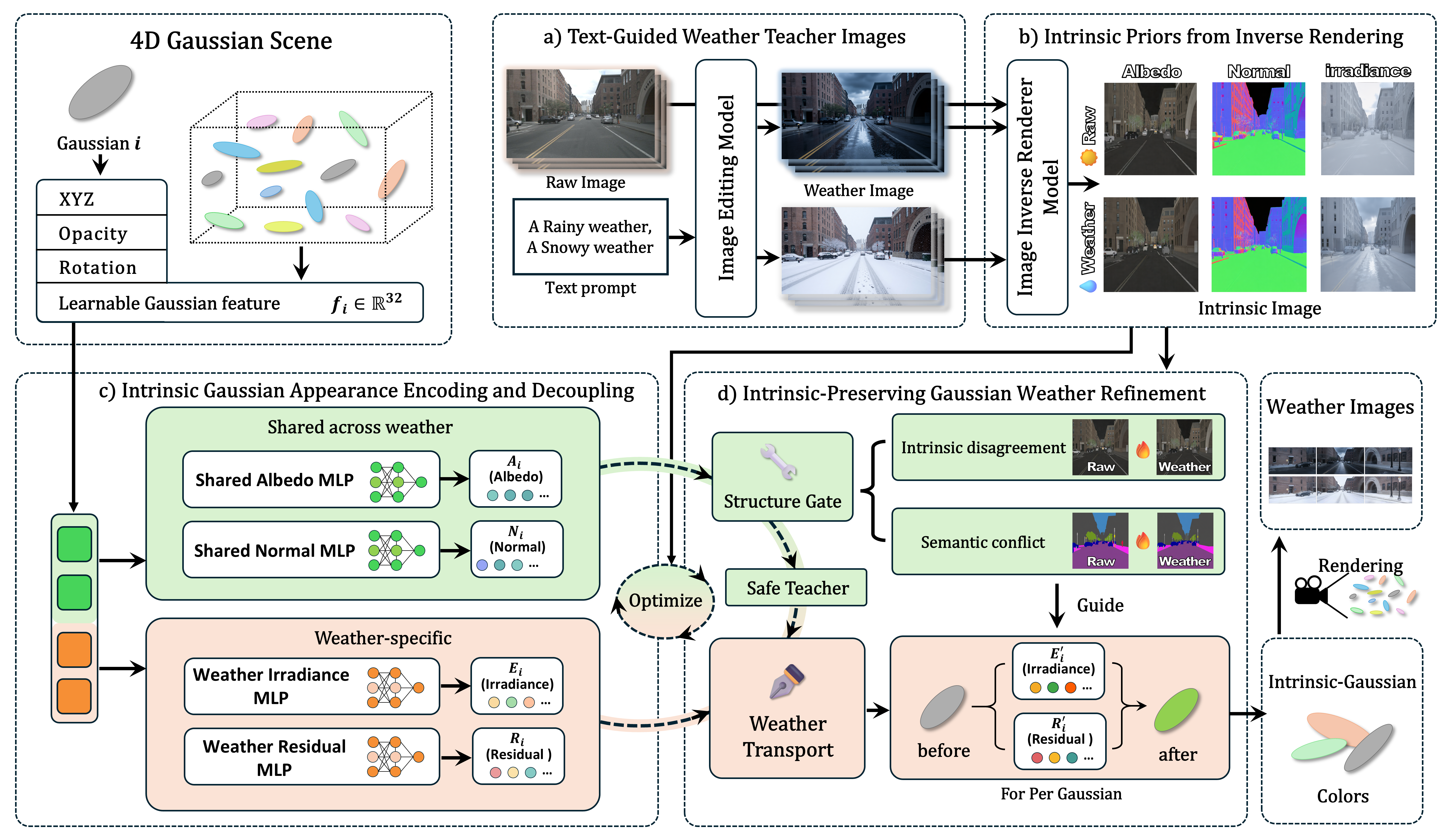}
    \caption{Overview of IGSW. (a) Qwen-Image produces target-weather teachers from the original driving observations. (b) A diffusion-based inverse renderer estimates albedo, normals, and irradiance for both original and teacher images. (c) Stage~I constructs an intrinsic-decoupled 4D Gaussian scene with weather-shared albedo/normals and weather-specific irradiance/residuals. (d) Stage~II uses a structure-preservation gate and per-Gaussian weather transport to suppress teacher-induced structural errors and strengthen target-weather illumination and atmosphere.}
    \label{fig:pipeline}
\end{figure*}

Given calibrated multi-view driving images with temporal correspondence, IGSW learns a dynamic Gaussian scene that renders a target weather condition while preserving the scene structure. Figure~\ref{fig:pipeline} summarizes the pipeline.

\subsection{Text-Guided Weather Teacher Images}
\label{sec:teachers}
We use Qwen-Image with weather-specific prompts that describe illumination, reflections, and environmental appearance for rain or snow while requesting preservation of the scene layout and visible content. Although the resulting teachers include weather-dependent changes to roads, buildings, and other surfaces, Qwen-Image edits frames independently and imposes no explicit cross-frame or cross-view constraint. Its sequences may therefore contain illumination drift, texture changes, or content errors such as clouds covering buildings.

Let $I_{t,v}^{o}\in[0,1]^{H\times W\times3}$ denote an original driving image, where $t$ and $v$ index time and camera view, $o$ denotes the original condition, and $H$ and $W$ are image height and width. Given a target-weather prompt $p_w$, the Qwen-Image editor $\mathcal{E}_{\mathrm{Qwen}}$ produces
\begin{equation}
T_{t,v}^{w}=\mathcal{E}_{\mathrm{Qwen}}(I_{t,v}^{o},p_w),
\qquad w\in\{\mathrm{rain},\mathrm{snow}\}.
\label{eq:teacher}
\end{equation}
We regard $T_{t,v}^{w}$ as an appearance reference for the target weather during subsequent Gaussian scene optimization, rather than geometrically and temporally consistent ground truth.

\subsection{Intrinsic Priors from Inverse Rendering}
\label{sec:intrinsic_prior}
We adopt the Stable Diffusion~3.5-based inverse renderer from IntrinsicWeather~\cite{r19} to decompose each original image and its Qwen teacher. The renderer estimates an albedo map, a normal map, and an irradiance map, which describe base color and material appearance, geometric structure, and illumination distribution, respectively. Let $\mathcal{D}_{\mathrm{SD}}$ denote this inverse renderer. Omitting time and camera subscripts for clarity,
\begin{equation}
\begin{aligned}
\mathcal{D}_{\mathrm{SD}}(I^{o})
    &=\bigl(\widetilde{A}^{o},\widetilde{N}^{o},\widetilde{E}^{o}\bigr),\\
\mathcal{D}_{\mathrm{SD}}(T^{w})
    &=\bigl(\widetilde{A}^{w},\widetilde{N}^{w},\widetilde{E}^{w}\bigr).
\end{aligned}
\label{eq:decomposition}
\end{equation}
Here $A$, $N$, and $E$ denote albedo, normals, and irradiance; superscripts $o$ and $w$ distinguish original and target-weather images; and tildes denote estimates from the diffusion model at each timestamp and viewpoint. The original albedo $\widetilde A^o$ and normals $\widetilde N^o$ provide references for material and structure that should remain stable across weather. The irradiance pair $\widetilde E^o,\widetilde E^w$ describes the desired change in illumination and atmosphere. Because these frame-wise two-dimensional estimates do not themselves form a consistent scene, we introduce their factorization into Gaussian appearance.

\begin{figure*}[!t]
\centering
\includegraphics[width=0.98\textwidth]{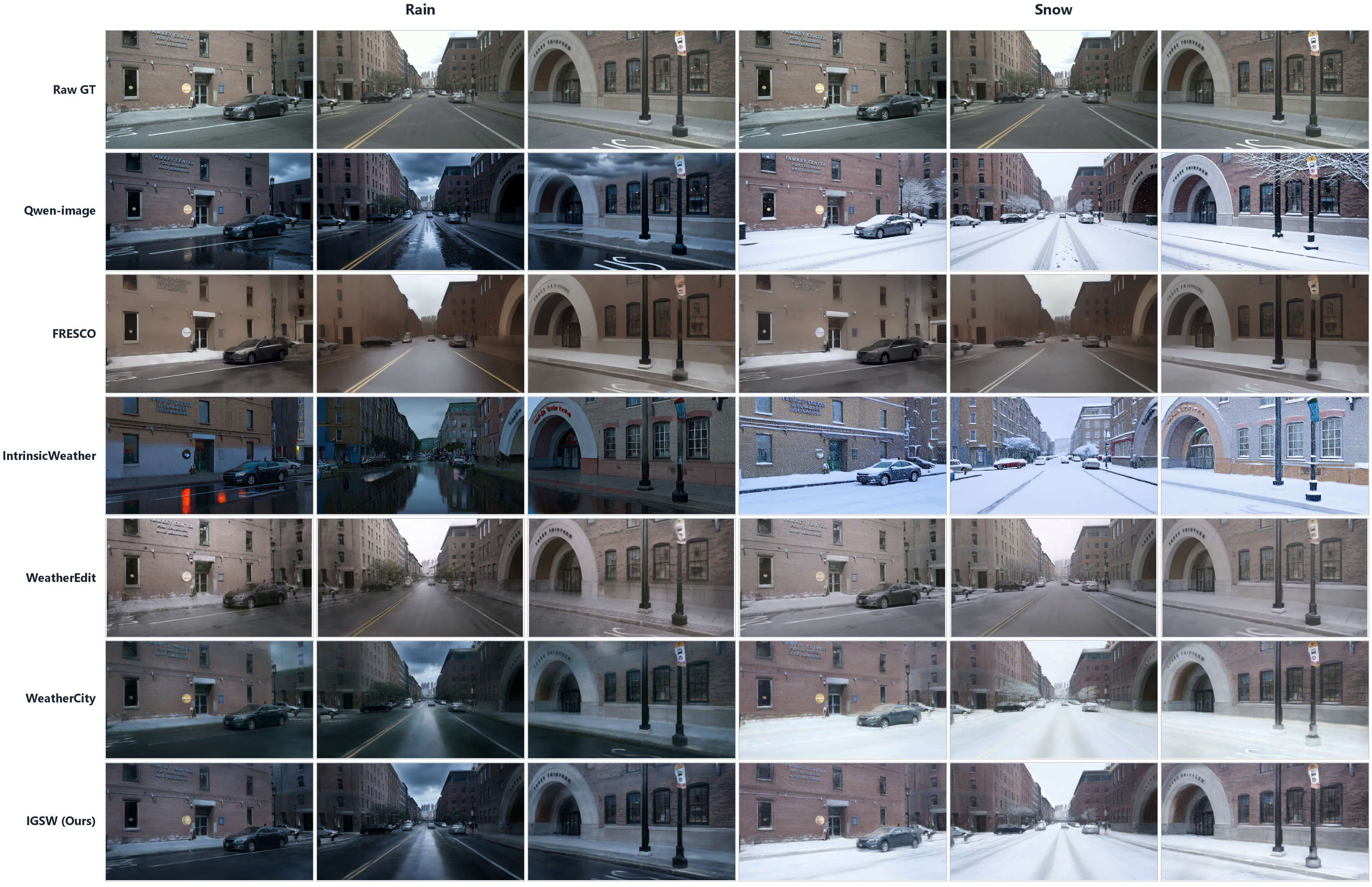}
\caption{Qualitative rain and snow comparisons on nuScenes. Rows show the original observations, Qwen-Image, FRESCO, IntrinsicWeather, WeatherEdit, WeatherCity, and IGSW. Columns use matched camera views and weather conditions.}
\label{fig:qual_nuscenes}
\end{figure*}

\begin{table*}[!t]
\centering
\caption{Complete-sequence quantitative comparison on nuScenes. Results are averaged equally across ten scenes; higher is better for CLIP-S, CLIP-DS, and Sem-CS, while lower is better for $\warp$. Best values are bold.}
\label{tab:nuscenes}
\footnotesize
\setlength{\tabcolsep}{4.4pt}
\renewcommand{\arraystretch}{1.15}
\begin{tabular*}{\textwidth}{@{\extracolsep{\fill}}lcccccccc@{}}
\toprule
\multirow{2}{*}{Method} & \multicolumn{4}{c}{Rain} & \multicolumn{4}{c}{Snow} \\
\cmidrule(lr){2-5}\cmidrule(lr){6-9}
& CLIP-S$\uparrow$ & CLIP-DS$\uparrow$ & Sem-CS$\uparrow$ & $\warp\downarrow$
& CLIP-S$\uparrow$ & CLIP-DS$\uparrow$ & Sem-CS$\uparrow$ & $\warp\downarrow$ \\
\midrule
Qwen-Image & 0.8248 & \textbf{0.2090} & 0.6748 & 0.0294 & 0.7648 & 0.2692 & 0.6253 & 0.0516 \\
FRESCO & 0.8070 & 0.0543 & 0.5693 & 0.0174 & 0.7885 & 0.0897 & 0.5443 & 0.0172 \\
IntrinsicWeather & 0.6682 & 0.1380 & 0.5497 & 0.0379 & 0.6437 & 0.1563 & 0.4969 & 0.0529 \\
WeatherEdit & 0.8443 & 0.1122 & 0.6863 & 0.0148 & 0.8269 & 0.1412 & 0.6839 & 0.0127 \\
WeatherCity & 0.8582 & 0.1289 & 0.7129 & 0.0023 & 0.8652 & 0.1876 & 0.6772 & 0.0038 \\
\midrule
IGSW (Ours) & \textbf{0.8718} & 0.1599 & \textbf{0.8992} & \textbf{0.0021}
& \textbf{0.8782} & \textbf{0.2732} & \textbf{0.8792} & \textbf{0.0028} \\
\bottomrule
\end{tabular*}
\end{table*}

\subsection{Intrinsic Gaussian Appearance Encoding and Decoupling}
\label{sec:intrinsic_gaussian}
Stage~I lifts the two-dimensional intrinsic factors into a 4D Gaussian scene. It preserves the Gaussian geometric representation while decomposing background appearance into weather-shared albedo and normals and weather-specific irradiance and residuals. Each intrinsic-decoupled Gaussian $g_i$ is represented as
\begin{equation}
g_i=(\mu_i,s_i,q_i,\alpha_i,f_i),
\quad
\begin{aligned}
\mu_i&\in\R^3, & s_i&\in\R^3,\\
q_i&\in\R^4, & \alpha_i&\in[0,1],\\
f_i&\in\R^{32}.&&
\end{aligned}
\label{eq:gaussian}
\end{equation}
Here $\mu_i$, $s_i$, $q_i$, and $\alpha_i$ denote position, scale, rotation quaternion, and opacity, respectively, while $f_i$ is a learned 32-dimensional appearance feature.

\subsubsection{Weather-Shared Albedo and Normals}
To retain a common material and structural representation across conditions, we decode albedo and normals through weather-shared heads. For background Gaussian $i$,
\begin{equation}
A_i=\sigma\!\left(h_A(f_i)\right),
\qquad
N_i=\tanh\!\left(h_N(f_i)\right),
\label{eq:shared_heads}
\end{equation}
where $A_i\in[0,1]^3$ is albedo, $N_i\in[-1,1]^3$ is a normal-direction encoding, and $\sigma$ and $\tanh$ are sigmoid and hyperbolic-tangent activations. Differentiable rasterization produces corresponding two-dimensional intrinsic maps, which receive feature-level supervision from the original-image albedo and normals in Sec.~\ref{sec:objectives}. Neither head receives a weather condition; the same Gaussian therefore uses the same albedo and normal representation for original and target weather.

\subsubsection{Weather-Specific Irradiance and Residuals}
We therefore assign separate irradiance and residual heads to the original and target-weather conditions. Irradiance models lighting, shadows, and ambient brightness, while the residual compensates for effects that cannot be expressed by albedo--irradiance multiplication alone. For condition $x\in\{o,w\}$,
\begin{equation}
E_i^x=\sigma\!\left(h_E^x(f_i)\right),
\qquad
R_i^x=\tanh\!\left(h_R^x(f_i)\right),
\label{eq:weather_heads}
\end{equation}
where $h_E^x$ and $h_R^x$ are condition-specific decoders, and $E_i^x$ and $R_i^x$ are three-channel irradiance and RGB residuals. The decoders differ across conditions but share the same Gaussian feature $f_i$. The color of Gaussian $i$ under condition $x$ is
\begin{equation}
c_i^x=\clip_{[0,1]}\!\left(A_i\odot E_i^x+\lambda_R R_i^x\right),
\label{eq:color}
\end{equation}
where $\odot$ denotes channel-wise multiplication and $\clip_{[0,1]}$ restricts RGB values to the valid range.

\subsection{Intrinsic-Preserving Gaussian Weather Refinement}
\label{sec:refinement}
Stage~I creates a unified scene representation, but joint optimization can still smooth weather texture, underfit local teacher appearance, or reproduce erroneous teacher content. A structure-preservation gate modifies supervision according to teacher reliability, and weather transport parameters attached to the Gaussians adapt illumination and residual appearance.

\subsubsection{Semantic- and Intrinsic-Consistency Structure Gate}
A soft gate identifies teacher regions where stable scene content requires protection.

For each aligned original--teacher pair, UPerNet--ConvNeXt-XL predicts semantic maps $S^o$ and $S^w$. Let $P(p)$ indicate whether pixel $p$ in the original image belongs to a protected category: wall, building, road, sidewalk, person, vehicle, house, or skyscraper. A category change in the teacher defines the semantic-conflict mask
\begin{equation}
M_s(p)=P(p)\ind\!\left[S^o(p)\neq S^w(p)\right].
\label{eq:semantic_conflict}
\end{equation}
Semantic disagreement alone does not cover every local structural defect. We therefore compare the albedo and normal maps from Sec.~\ref{sec:intrinsic_prior}. Let $\widetilde n$ denote a normal direction obtained by mapping the estimated normal map from $[0,1]$ to $[-1,1]$ and then normalizing it. The intrinsic discrepancy is
\begin{equation}
D(p)=\frac{1}{2}\left[
\frac{\nrm{\widetilde A^o(p)-\widetilde A^w(p)}_1}{3}
+\frac{1-\widetilde n^o(p)^\top\widetilde n^w(p)}{2}
\right].
\label{eq:intrinsic_discrepancy}
\end{equation}
Because inverse-rendered intrinsic maps are themselves uncertain, a large discrepancy is not treated as a structural error everywhere. It is used only near semantic inconsistencies to assess teacher reliability. Let $\mathcal D_r$ denote square max-pooling dilation with radius $r$, and let $\mathcal P_r$ denote average pooling under the same convention. We define
\begin{equation}
\begin{aligned}
M_i={}&P\odot\mathcal D_4\!\left(\ind[S^o\neq S^w]\right)
        \odot\ind[D>\tau],\\
G={}&\mathcal P_9\!\left(\mathcal D_5(M_s\vee M_i)\right).
\end{aligned}
\label{eq:structure_gate}
\end{equation}
The resulting $G\in[0,1]$ is the structure-preservation gate; larger values indicate stronger protection.

Using the original image directly in protected regions would also restore original lighting and weaken the desired weather appearance. We instead preserve the spatial content of the original image while transferring teacher color statistics. In reliable pixels that are semantically consistent and satisfy $D\leq\tau$, we compute the per-channel mean and standard deviation of original and teacher RGB values for each semantic category. The resulting mapping is
\begin{equation}
\begin{aligned}
\mathcal S_w(I^o)(p)=\clip_{[0,1]}\!\Bigg[
&\bigl(I^o(p)-\mu_o\bigr)\odot\\[-2pt]
&\clip_{[0.65,1.35]}\!\left(\frac{\sigma_w}{\sigma_o}\right)+\mu_w
\Bigg].
\end{aligned}
\label{eq:style_mapping}
\end{equation}
The statistics are taken from the category containing $p$; if that category has fewer than 256 reliable pixels, global reliable-region statistics are used. This mapping changes color and contrast without moving pixels. The safe teacher is then
\begin{equation}
T_{\mathrm{safe}}^w=(1-G)\odot T^w
+G\odot\mathcal S_w(I^o).
\label{eq:safe_teacher}
\end{equation}
When $G$ is small, supervision comes primarily from the Qwen weather teacher. When $G$ is large, it comes from weather-styled original content and is supplemented by an original-image edge constraint. The gate therefore reduces the transfer of structurally unreliable teacher content while retaining target-weather tone.

\subsubsection{Per-Gaussian Weather Transport}
The structure gate determines which teacher content is reliable. Weather transport controls how target-weather appearance changes over a fixed scene structure. We attach two target-weather scalars $u_i^w,v_i^w\in\R$ to each background Gaussian and use them to modulate irradiance and residuals:
\begin{equation}
\begin{aligned}
E_i^{w'}&=E_i^w\exp\!\left(2\tanh(u_i^w)\right),\\
R_i^{w'}&=R_i^w+2\tanh(v_i^w)\ind_3.
\end{aligned}
\label{eq:transport}
\end{equation}
Here $u_i^w$ controls a multiplicative illumination gain, $v_i^w$ controls an additive residual shift, and $\ind_3$ broadcasts one scalar to all RGB channels. Both variables are initialized to zero, so the transport initially preserves the Stage~I appearance. The hyperbolic tangent bounds the gain to $[e^{-2},e^2]$ and the residual increment to $[-2,2]$.

Unlike independent parameters for each color channel, these scalars form a low-dimensional achromatic path for adapting illumination intensity and residual response at the individual Gaussian level. The refined color is
\begin{equation}
c_i^{w'}=\clip_{[0,1]}\!\left(A_i\odot E_i^{w'}
+\lambda_R R_i^{w'}\right).
\label{eq:refined_color}
\end{equation}
Stage~II retains the Stage~I differentiable renderer and optimizes weather-related parameters with the safe teacher and structure-preserving constraints. Shared albedo and normals remain fixed. Because the transport parameters are attached to scene Gaussians, views observing the same primitive share its weather adaptation rather than learning image-specific corrections.

\begin{figure*}[!t]
\centering
\includegraphics[width=0.98\textwidth]{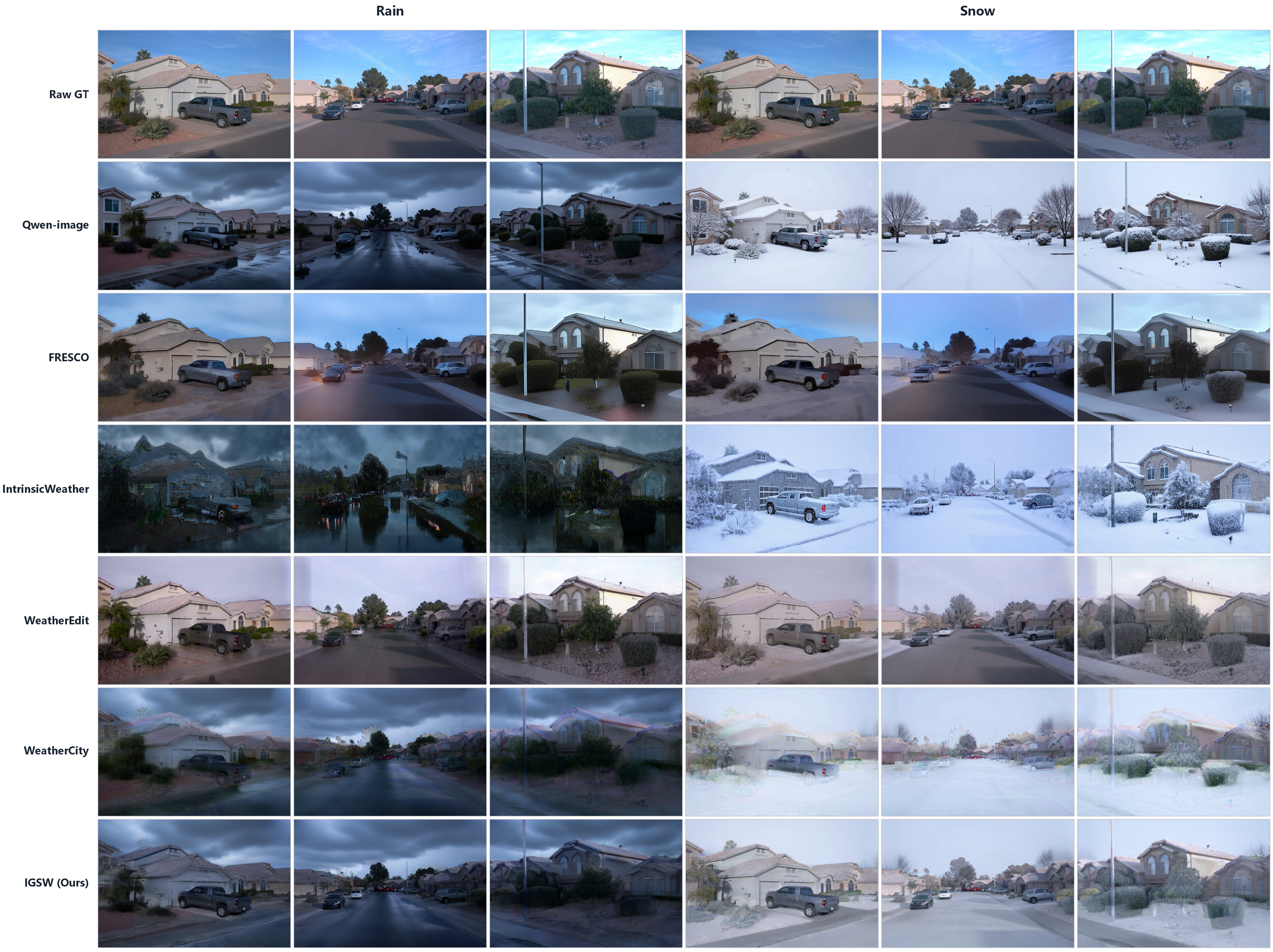}
\caption{Qualitative rain and snow comparisons on Waymo. The method order matches Fig.~\ref{fig:qual_nuscenes}; viewpoints and weather conditions are aligned across methods.}
\label{fig:qual_waymo}
\end{figure*}

\subsection{Training Objectives and Final Rendering}
\label{sec:objectives}

\subsubsection{Intrinsic Feature Constraints}
Intrinsic maps estimated by a diffusion model may contain local misalignment and numerical errors. We therefore compare local features rather than treating them as pixel-accurate ground truth. A fixed lightweight encoder $\Phi$ extracts low-frequency color with average pooling and Sobel edge magnitudes, concatenates both feature types, and normalizes each patch representation.

\textbf{Weather-shared intrinsic constraint.}
For both original and target-weather renders, original-image albedo and normals provide the reference:
\begin{equation}
\mathcal L_{\mathrm{AN}}^x
=\frac{1}{2}\sum_{Z\in\{A,N\}}
d_{\cos}\!\left(\Phi(\widehat Z^x),\Phi(\widetilde Z^o)\right),
\quad x\in\{o,w\}.
\label{eq:an_loss}
\end{equation}
Hats denote rendered intrinsic maps, while tildes denote external estimates. Normal maps are converted to the reference range before encoding. The distance $d_{\cos}$ is a patch-wise cosine distance weighted by valid-region coverage.

\textbf{Weather-specific intrinsic constraint.}
For the target weather, we constrain the direction of the irradiance change rather than requiring pixel-wise agreement:
\begin{equation}
\mathcal L_{\mathrm{dir}}=d_{\cos}\!\left(
\Phi(\widehat E^w)-\Phi(\widetilde E^o),
\Phi(\widetilde E^w)-\Phi(\widetilde E^o)
\right).
\label{eq:direction_loss}
\end{equation}
Each patch difference is normalized before computing cosine distance. Both directions start from the original external irradiance estimate, encouraging the predicted illumination to follow the teacher's weather-dependent trend.

\subsubsection{Stage I Objective}
Stage~I jointly optimizes scene geometry, the shared code, and all decoder branches. A frozen VGG-19 content loss $\mathcal L_{\mathrm{content}}$ anchors the original observations during weather fitting. Sparse-depth supervision, foreground-occupancy mask supervision, camera color-correction regularization, Gaussian-shape regularization, and dynamic-object regularization are grouped into $\mathcal L_{\mathrm{aux}}$. The complete Stage~I objective is
\begin{equation}
\mathcal L_{\mathrm{I}}
=\sum_{x\in\{o,w\}}\left(
\mathcal L_{\mathrm{rec}}^x
+\mathcal L_{\mathrm{content}}^x
+\lambda_{\mathrm{AN}}\mathcal L_{\mathrm{AN}}^x
+\mathcal L_{\mathrm{aux}}^x\right)
+\lambda_{\mathrm{dir}}\mathcal L_{\mathrm{dir}}.
\label{eq:stage1}
\end{equation}
\subsubsection{Stage II Objective}
Stage~II fixes geometry and shared intrinsic representations and uses $T_{\mathrm{safe}}^w$ from Sec.~\ref{sec:refinement} to refine target-weather appearance. The weather reconstruction loss $\mathcal L_{\mathrm{weather}}$ combines full-resolution and low-frequency $\ell_1$ errors to constrain local appearance and global weather tone. A local VGG feature-matching loss $\mathcal L_{\mathrm{local}}$ searches within a bounded neighborhood to tolerate small teacher misalignments. In regions selected by $G$, an original-image Sobel loss $\mathcal L_{\mathrm{edge}}$ protects walls, windows, and object contours. Finally, $\mathcal L_{uv}$ penalizes the squared magnitudes of the transport variables. The refinement objective is
\begin{equation}
\begin{aligned}
\mathcal L_{\mathrm{II}}={}&\mathcal L_{\mathrm{weather}}
+\lambda_{\mathrm{local}}\mathcal L_{\mathrm{local}}
+\lambda_{\mathrm{edge}}\mathcal L_{\mathrm{edge}}\\
&+\lambda_{\mathrm{dir}}\mathcal L_{\mathrm{dir}}
+\lambda_{uv}\mathcal L_{uv}.
\end{aligned}
\label{eq:stage2}
\end{equation}
Additional implementation details are provided in the supplementary material.

\subsubsection{Final Weather-Scene Rendering}
Given time $t$, camera $k$, and weather $w$, the model first determines each Gaussian's position and shape and computes its condition-specific color. A background Gaussian uses the refined irradiance and residual from Sec.~\ref{sec:refinement} together with shared albedo; dynamic objects use their corresponding weather appearance branch.

Each Gaussian is transformed into camera coordinates and projected onto the image plane using the calibrated camera model. Its two-dimensional covariance is
\begin{equation}
\Sigma_{i,t,k}^{2\mathrm D}
=J_{i,t,k}W_k\Sigma_{i,t}W_k^\top J_{i,t,k}^\top,
\label{eq:covariance}
\end{equation}
where $\Sigma_{i,t}$ is the three-dimensional covariance at time $t$, determined by scale and rotation; $W_k$ is the rotational component of the world-to-camera transformation; and $J_{i,t,k}$ is the Jacobian of perspective projection at the Gaussian center.

For pixel $p$, contributing Gaussians are sorted from near to far and alpha-composited. The final target-weather image is
\begin{equation}
\begin{aligned}
\widehat I_{t,k}^{w}(p)=\mathcal A_k\!\Bigg[&
\sum_{i=1}^{N}c_i^{w,*}a_i(p)
\prod_{j<i}\bigl(1-a_j(p)\bigr)\\
&+B_{t,k}^{w}(p)\prod_{i=1}^{N}\bigl(1-a_i(p)\bigr)
\Bigg].
\end{aligned}
\label{eq:rendering}
\end{equation}
Here $c_i^{w,*}$ is the optimized Gaussian color, and $a_i(p)$ is its effective opacity at $p$, determined by primitive opacity and the projected Gaussian distribution. The product over preceding Gaussians is accumulated transmittance and accounts for occlusion. $B_{t,k}^{w}(p)$ is the target-weather sky color, and $\mathcal A_k$ is camera-dependent affine color correction.

\begin{table*}[!t]
\centering
\caption{Complete-sequence quantitative comparison on the Waymo Open Dataset. Metric directions match Table~\ref{tab:nuscenes}; best values are bold.}
\label{tab:waymo}
\footnotesize
\setlength{\tabcolsep}{4.4pt}
\renewcommand{\arraystretch}{1.15}
\begin{tabular*}{\textwidth}{@{\extracolsep{\fill}}lcccccccc@{}}
\toprule
\multirow{2}{*}{Method} & \multicolumn{4}{c}{Rain} & \multicolumn{4}{c}{Snow} \\
\cmidrule(lr){2-5}\cmidrule(lr){6-9}
& CLIP-S$\uparrow$ & CLIP-DS$\uparrow$ & Sem-CS$\uparrow$ & $\warp\downarrow$
& CLIP-S$\uparrow$ & CLIP-DS$\uparrow$ & Sem-CS$\uparrow$ & $\warp\downarrow$ \\
\midrule
Qwen-Image & 0.8450 & 0.1942 & 0.5099 & 0.0547 & 0.7971 & 0.2007 & 0.4571 & 0.0672 \\
FRESCO & 0.8703 & 0.1487 & 0.6376 & 0.0195 & 0.8258 & 0.1557 & 0.6710 & 0.0184 \\
IntrinsicWeather & 0.6046 & 0.1601 & 0.3919 & 0.0224 & 0.7317 & 0.1929 & 0.5676 & 0.0273 \\
WeatherEdit & 0.9015 & 0.2095 & 0.8036 & 0.0118 & 0.8532 & 0.2055 & 0.8025 & 0.0106 \\
WeatherCity & 0.7773 & 0.1375 & 0.7398 & 0.0090 & 0.8002 & 0.0937 & 0.6633 & 0.0101 \\
\midrule
IGSW (Ours) & \textbf{0.9031} & \textbf{0.2940} & \textbf{0.8723} & \textbf{0.0009}
& \textbf{0.8971} & \textbf{0.2869} & \textbf{0.8050} & \textbf{0.0019} \\
\bottomrule
\end{tabular*}
\end{table*}

\section{Experiments}
\label{sec:experiments}

\subsection{Experimental Setup}
\textbf{Data and training.}
We evaluate rain and snow editing on ten scenes from nuScenes~\cite{r25} and on the Waymo Open Dataset~\cite{r26}. Each scene contains 100 synchronized timestamps from three front-facing cameras, yielding 300 images per weather condition. Qwen-Image produces the weather teachers, and the Stable Diffusion~3.5-based model in IntrinsicWeather produces the intrinsic decompositions. Separate original/rain and original/snow scene models are trained. The raw nuScenes and Waymo images are $1600\times900$ and $1920\times1280$. Stage~I runs for 30,000 steps. Stage~II contains 800 original-observation recovery steps followed by 3,000 weather-refinement steps. All experiments are conducted on a server with an NVIDIA RTX~4080 GPU. Optimizer settings, loss weights, and gate construction are detailed in the supplementary material.

\textbf{Evaluation metrics.}
We use CLIP-S, CLIP-DS, Sem-CS, and $\warp$. CLIP-S measures image-feature similarity to the original image, while CLIP-DS measures alignment between the original-to-weather image change and a fixed text-feature direction. Both use CLIP ViT-B/32~\cite{r27}. Sem-CS computes frequency-weighted semantic IoU with ADE20K-pretrained UPerNet--ConvNeXt-XL~\cite{r29,r30}. $\warp$ uses RAFT-Large~\cite{teed2020raft} flow estimated from adjacent original frames to align the corresponding weather frames and then computes squared RGB error. Every method reuses the same flow and valid masks. Higher CLIP-S, CLIP-DS, and Sem-CS are better, whereas lower $\warp$ is better. Image-wise metrics use all 300 images; temporal metrics use all 297 within-camera adjacent pairs. The nuScenes results equally average the ten scenes. Complete definitions are given in Supplementary Sec.~S2.

\textbf{Compared methods.}
We compare with the image editor Qwen-Image, the video editor FRESCO, the intrinsic inverse/forward rendering method IntrinsicWeather, and the 4D scene-editing methods WeatherEdit and WeatherCity. Scene-level methods based on edited supervision use the same text-guided weather teachers for a fair comparison. Table~\ref{tab:methods} summarizes their method categories and representational capabilities.

\begin{figure*}[!t]
\centering
\includegraphics[width=0.98\textwidth]{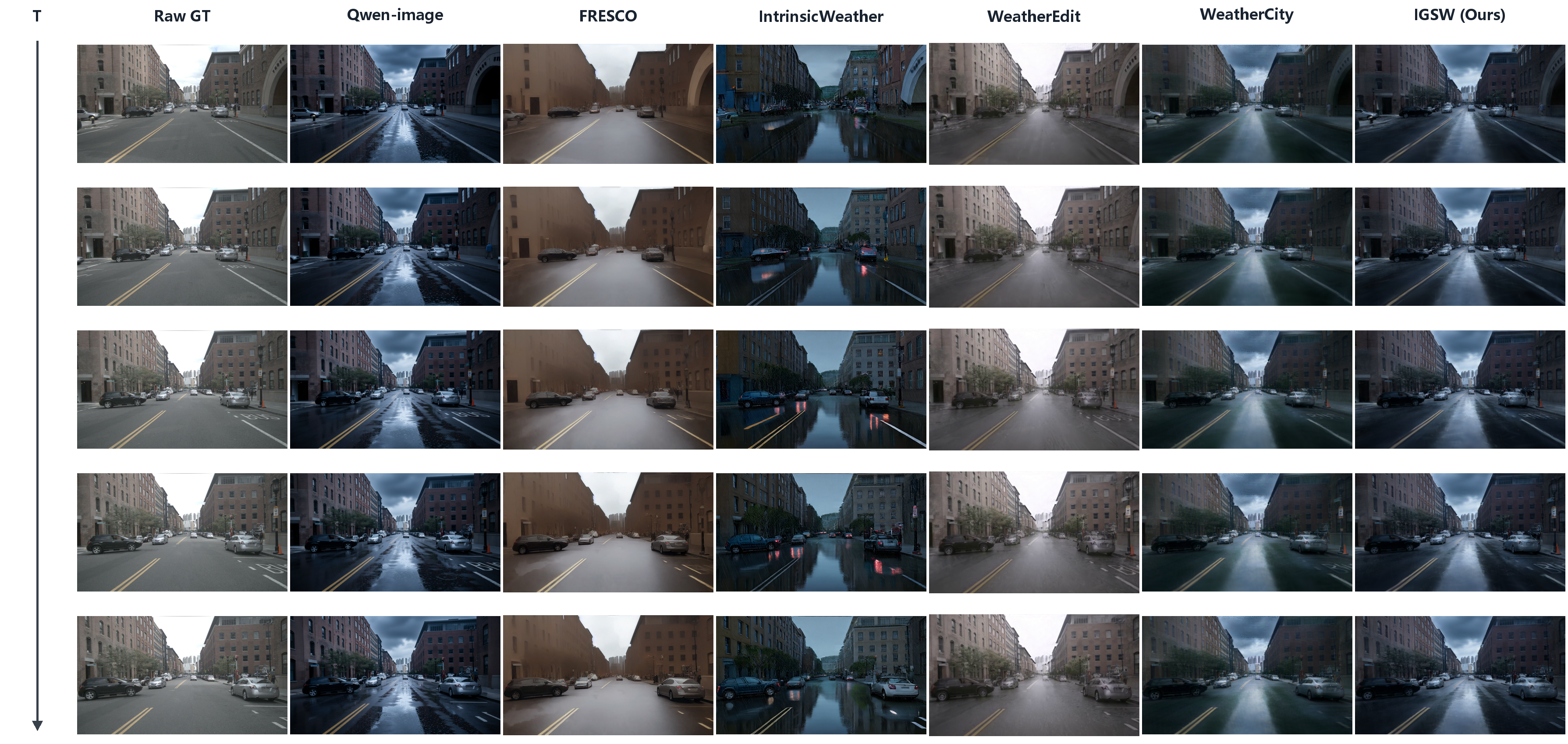}
\caption{Temporal rain comparison on nuScenes at frames 49, 54, 59, 64, and 69. The samples are fixed across methods; quantitative evaluation uses all 297 adjacent frame pairs. The nuScenes snow and Waymo sequences are provided in the supplementary material.}
\label{fig:temporal_nusc}
\end{figure*}

\begin{table}[!t]
\centering
\caption{Temporal warping error on nuScenes. Lower is better.}
\label{tab:temporal}
\small
\renewcommand{\arraystretch}{1.12}
\begin{tabular*}{\columnwidth}{@{\extracolsep{\fill}}lcc@{}}
\toprule
Method & Rain $\warp\downarrow$ & Snow $\warp\downarrow$ \\
\midrule
Qwen-Image & 0.0294 & 0.0516 \\
FRESCO & 0.0174 & 0.0172 \\
IntrinsicWeather & 0.0379 & 0.0529 \\
WeatherEdit & 0.0148 & 0.0127 \\
WeatherCity & 0.0023 & 0.0038 \\
\midrule
IGSW (Ours) & \textbf{0.0021} & \textbf{0.0028} \\
\bottomrule
\end{tabular*}
\end{table}

\begin{figure}[!t]
\centering
\includegraphics[width=\columnwidth]{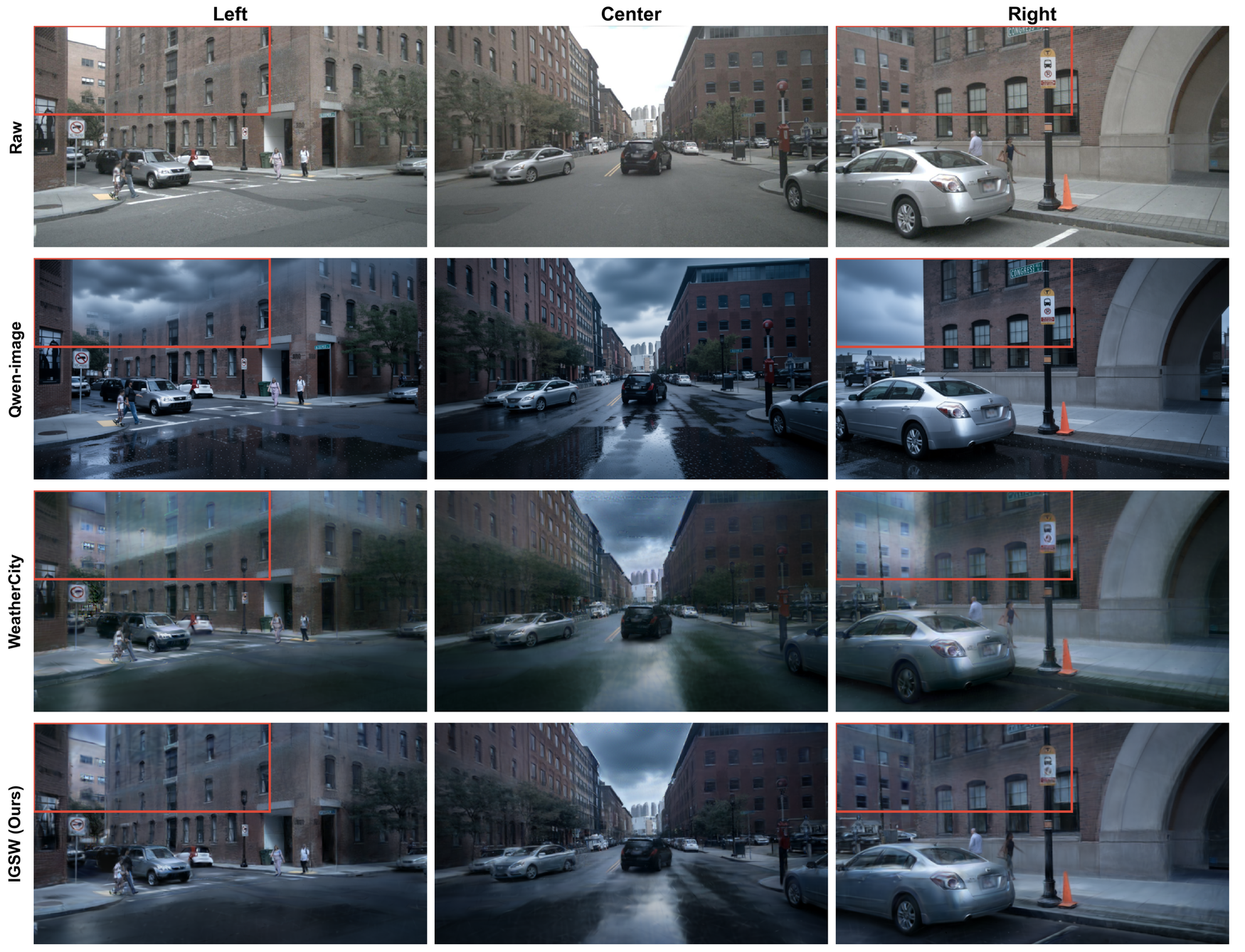}
\caption{Structural comparison at one timestamp and three views. Rows show the original observations, Qwen-Image, WeatherCity, and IGSW. Clouds in the teacher incorrectly cover visible buildings; IGSW reduces this error while adapting the scene to rainy appearance.}
\label{fig:structure}
\end{figure}

\subsection{Quantitative and Qualitative Results}
Tables~\ref{tab:nuscenes} and~\ref{tab:waymo} report complete-sequence results. IGSW achieves the highest CLIP-S and Sem-CS and the lowest $\warp$ in all four dataset--weather settings. It also obtains the highest CLIP-DS in three of the four settings.

On nuScenes, Qwen-Image attains the highest rain CLIP-DS (0.2090), indicating a stronger match to the prescribed weather direction under that metric. Its substantially larger temporal error and lower semantic consistency show that weather-direction alignment alone does not characterize scene preservation.

Figures~\ref{fig:qual_nuscenes} and~\ref{fig:qual_waymo} compare matched views. In the displayed cases, FRESCO and WeatherEdit produce relatively weak weather changes. IntrinsicWeather exhibits irregular noise and stripe artifacts together with local structural degradation. WeatherCity provides scene-level rendering but can inherit structural errors from the Qwen teachers and remains farther from their illumination and atmosphere. IGSW more closely approaches the teacher's weather-dependent lighting and color while retaining building, road, vehicle, and pedestrian structure.

\begin{table*}[!t]
\centering
\caption{Module ablations using complete 100-timestamp, three-view evaluation. Best values are bold.}
\label{tab:modules}
\footnotesize
\setlength{\tabcolsep}{4.2pt}
\renewcommand{\arraystretch}{1.15}
\begin{tabular*}{\textwidth}{@{\extracolsep{\fill}}lcccccccc@{}}
\toprule
\multirow{2}{*}{Method} & \multicolumn{4}{c}{Rain} & \multicolumn{4}{c}{Snow} \\
\cmidrule(lr){2-5}\cmidrule(lr){6-9}
& CLIP-S$\uparrow$ & CLIP-DS$\uparrow$ & Sem-CS$\uparrow$ & $\warp\downarrow$
& CLIP-S$\uparrow$ & CLIP-DS$\uparrow$ & Sem-CS$\uparrow$ & $\warp\downarrow$ \\
\midrule
WeatherCity (baseline) & 0.858270 & 0.128999 & 0.712923 & 0.002399 & 0.865202 & 0.187677 & 0.677263 & 0.003811 \\
Stage~I only & 0.858933 & 0.132594 & 0.714228 & 0.002265 & 0.866956 & 0.188946 & 0.677516 & 0.003186 \\
Stage~II only & 0.866129 & 0.127921 & 0.898980 & \textbf{0.001632} & 0.864229 & 0.220130 & 0.877846 & \textbf{0.002207} \\
\midrule
IGSW (Stage~I + II) & \textbf{0.871860} & \textbf{0.159925} & \textbf{0.899250} & 0.002152
& \textbf{0.878254} & \textbf{0.273236} & \textbf{0.879220} & 0.002844 \\
\bottomrule
\end{tabular*}
\end{table*}

\begin{figure*}[!t]
\centering
\includegraphics[width=0.98\textwidth]{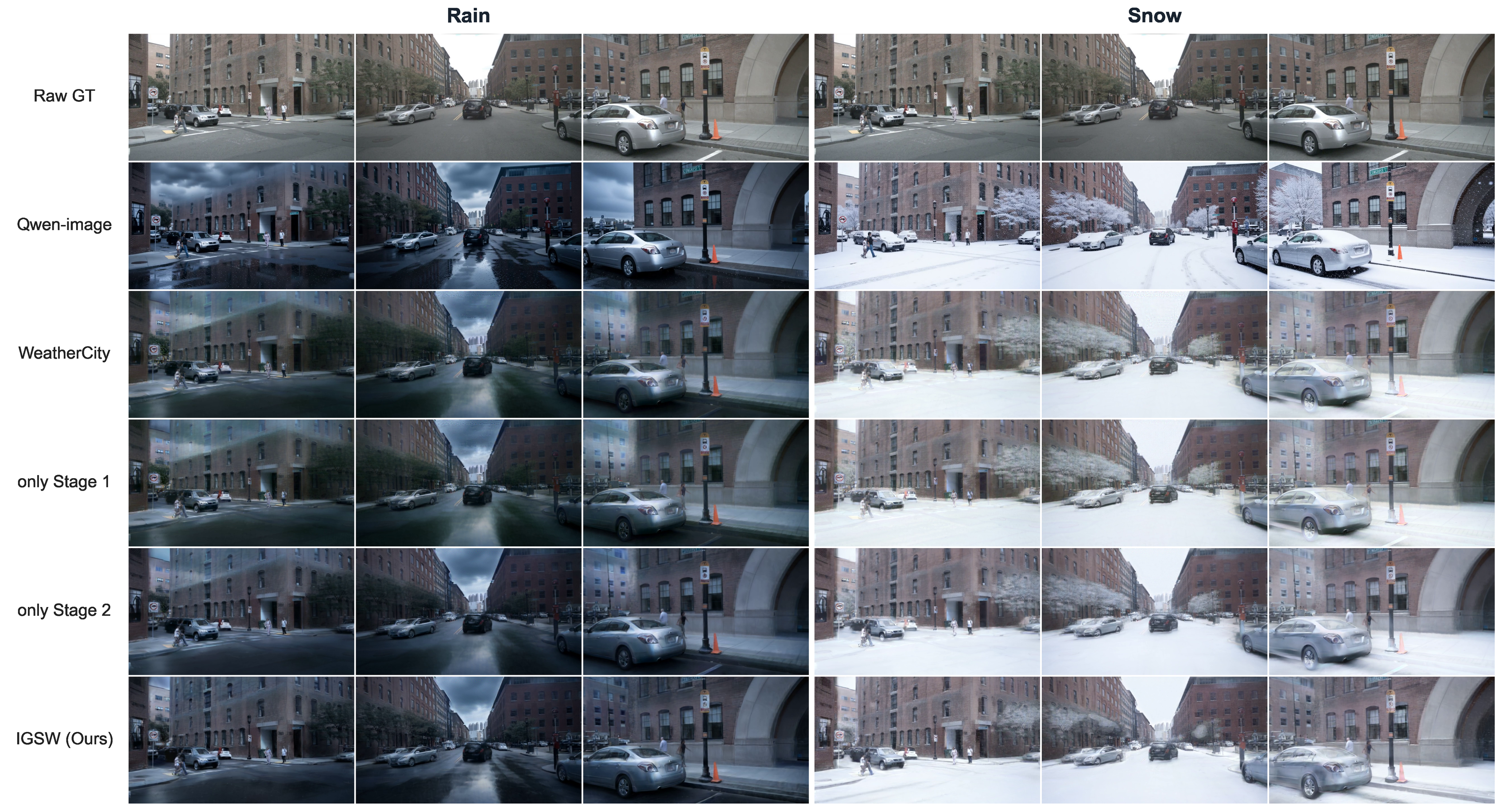}
\caption{Module ablations for rain (left) and snow (right). Rows show the original observations, Qwen teachers, WeatherCity, Stage~I only, Stage~II only, and complete IGSW. Each row contains three matched camera views.}
\label{fig:modules}
\end{figure*}

\subsection{Temporal and Structural Consistency}
\textbf{Temporal consistency.}
Table~\ref{tab:temporal} reports nuScenes temporal errors. IGSW improves upon WeatherCity under both weather conditions and reduces $\warp$ by more than one order of magnitude relative to the Qwen teachers. Figure~\ref{fig:temporal_nusc} shows rain appearance at five fixed timestamps. The intrinsic-decoupled Gaussian scene reduces frame-wise flicker and appearance jumps while retaining target-weather illumination and atmosphere. Supplementary Secs.~S3--S4 provide the corresponding nuScenes snow and Waymo rain/snow sequences.

\textbf{Structural consistency.}
Figure~\ref{fig:structure} isolates a teacher-induced structural error. Qwen-generated clouds cover parts of buildings that remain visible in the original observations, and WeatherCity reproduces part of this error. Through intrinsic decoupling and the Stage~II structure gate, IGSW substantially reduces the erroneous occlusion while maintaining rainy illumination.

\subsection{Ablation Studies}

\subsubsection{Module Ablations}
We evaluate the contribution of the two stages on complete 100-timestamp nuScenes sequences. The variants are WeatherCity as the baseline, Stage~I only, Stage~II only, and the complete two-stage model. Stage~I variants train for 30,000 steps. The Stage~II-only variant applies structure-preserving refinement without the intrinsic Stage~I representation, separating subsequent appearance optimization from intrinsic Gaussian modeling.

Table~\ref{tab:modules} shows that Stage~II only yields the lowest temporal error but trails the complete model on CLIP-S, CLIP-DS, and Sem-CS. Stage~I only modestly improves the baseline metrics, but without Stage~II it does not fully approach the illumination and atmosphere of the Qwen teachers. Combining both stages obtains the highest three similarity and semantic scores, with slightly higher temporal error than Stage~II only but lower error than the baseline. The quantitative trade-off and the visual comparisons in Fig.~\ref{fig:modules} support complementary roles for intrinsic factorization and protected appearance refinement.

Stage~I only remains close to WeatherCity and does not fully reproduce the teacher's weather-dependent lighting. Stage~II only transfers a stronger teacher style, but without the Stage~I factorization it can also reproduce structural errors. Complete IGSW combines the intrinsic representation with the gate and weather transport, improving both the quantitative balance and the displayed preservation of scene content.

\subsubsection{Component Ablations}
We further ablate shared albedo/normals, weather-specific irradiance/residuals, the structure-preservation gate, and per-Gaussian weather transport on nuScenes. The results reveal complementary roles in preserving content, expressing target weather, filtering unreliable teacher supervision, and adapting local appearance. Removing the gate weakens semantic consistency, while disabling weather transport reduces target-weather strength. Complete IGSW improves all four metrics over WeatherCity, with trade-offs between weather strength and temporal error across individual removals. The full table, visual comparisons, and analysis appear in Supplementary Sec.~S5 (Table~VII and Fig.~11).

\section{Conclusion}
IGSW provides a unified framework for weather-aware urban reconstruction with an intrinsic-decoupled Gaussian representation. It separates material and structural cues that should remain stable across weather from condition-dependent illumination and residual appearance. Structure-preserving refinement then combines reliability-aware teacher supervision with point-attached weather transport. Across the reported rain and snow experiments, this design improves the balance between target-weather expression, scene-content preservation, and temporal continuity, and reduces the structural errors introduced by image teachers in the displayed examples.

\end{document}